\documentclass[letterpaper, 10 pt, conference]{ieeeconf}  

\IEEEoverridecommandlockouts                              

\usepackage{graphics} 
\usepackage{subfigure}
\usepackage{polynom}
\usepackage{fixltx2e}
\usepackage{hyperref}
\usepackage{textgreek}
\usepackage{caption}
\usepackage{float}
\usepackage{siunitx}
\usepackage{epsfig} 
\usepackage[ruled]{algorithm2e} 
\usepackage{url}
\usepackage{amsmath} 

\title{\LARGE \bf
Autonomous Robot Navigation with Rich Information Mapping in Nuclear Storage Environments
}

\author{Maozhen Wang$^{1}$, Xianchao Long$^{2}$, Peng Chang$^{2}$ and Ta\c{s}k{\i}n~Pad{\i}r$^{2}$
\thanks{$^{1}$Mechanical and Industrial Engineering, Northeastern University, Boston, MA 02115, USA
       {\tt\small wang.mao@husky.neu.edu}}
\thanks{$^{2}$Electrical and Computer Engineering, Northeastern University, Boston, MA 02115, USA}}%

\begin{document}

\maketitle
\thispagestyle{empty}
\pagestyle{empty}

\begin{abstract}

This paper presents our approach to develop a method for an unmanned ground vehicle (UGV) to perform inspection tasks in nuclear environments using rich information maps. To reduce inspectors‘ exposure to elevated radiation levels, an autonomous navigation framework for the UGV has been developed to perform routine inspections such as counting containers, recording their ID tags and performing gamma measurements on some of them. In order to achieve autonomy, a rich information map is generated which includes not only the 2D global cost map consisting of obstacle locations for path planning, but also the location and orientation information for the objects of interest from the inspector's perspective. The UGV's autonomy framework utilizes this information to prioritize locations to navigate to perform the inspections. In this paper, we present our method of generating this rich information map, originally developed to meet the requirements of the International Atomic Energy Agency (IAEA) Robotics Challenge. We demonstrate the performance of our method in a simulated testbed environment containing uranium hexafluoride (UF6) storage container mock ups.

\end{abstract}

\section{INTRODUCTION}

There is an ever-increasing need to design and develop reliable robotics technologies in nuclear environments as elevated radiation levels pose dangerous conditions for humans whose jobs require them to be present in these places. In order to meet this need, many organizations are taking on initiatives to accelerate the development of autonomous robots for nuclear applications. For instance,  the International Atomic Energy Agency (IAEA) Robotics Challenge held in November 2017 was aimed at developing autonomous robots to conduct in-field inspections to ensure effective nuclear safeguards. As part of of IAEA's mission, inspectors routinely enter storage facilities to count containers, record their ID tags and perform gamma measurements. 2017 IAEA Robotics Challenge required participating teams to develop autonomous robots to perform these tasks in simulated environments.

For a robot to perform a given task in this scenario, it is an essential capability to develop autonomous navigation algorithms that rely heavily on the knowledge from the operating environment. There exist numerous methods on simultaneous localization and mapping (SLAM) to enable mobile robot navigation in unknown environments with various sensor configurations \cite{Labbé2017}, \cite{7946260}, \cite{KohlbrecherMeyerStrykKlingaufFlexibleSlamSystem2011}. However, in most SLAM-based navigation techniques, the generated map only consists of information about the free space accessible by the robot. In this paper, by unifying SLAM with object recognition, we propose a method that generates a rich information map which represents not only the free space for robot navigation but also the location and orientation information for objects of interest for the operator. This approach is also known as semantic SLAM \cite{8265293}.

   \begin{figure}[tpb]
      \centering
      \includegraphics[width=0.45\textwidth]{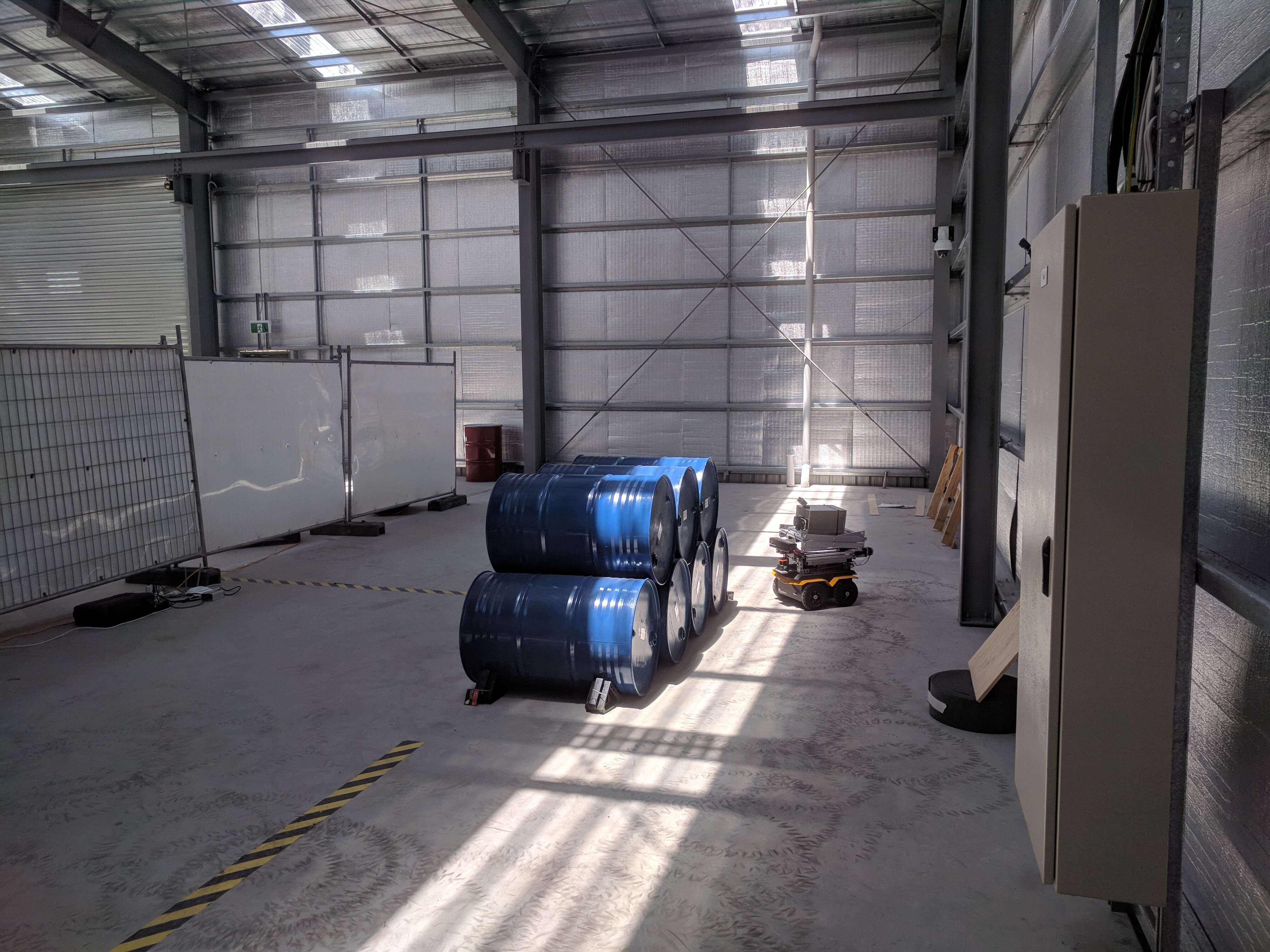}
      \caption{Demonstration of autonomous UGV at the IAEA Robotics Challenge.}
      \label{robot}
   \end{figure}

\renewcommand{\thefootnote}{\roman{footnote}}

\subsection{Problem Description}
The research and development effort presented in this paper is motivated by the IAEA Robotics Challenge, held in November 2017 in Brisbane, Australia. The objective of the challenge was to deploy an autonomous unmanned ground vehicle (UGV) to inspect Uranium Hexafluoride (UF6) storage cylinders, typically organized into piles (Figure \ref{robot}). IAEA UGV challenge requirements from the autonomous navigation perspective can be summarized as follow. (i) The robot must autonomously navigate in a 6~m by 10~m field which includes a pile of cylinders of size approximately 100~cm L x 50~cm D. (ii) The robot must identify and localize cylinders within the workspace. (iii) The robot must keep a count of the cylinders. 
In addition, it is required that the map is readable by the inspectors to allow for monitoring of the UF6 storage without entering the radioactive area. Detailed description of the challenge can be found at IAEA's website\footnote{www.iaea.org/es/topics/safeguards-in-practice/robotics-challenge-2017}. 

\subsection{System Description}

\begin{figure}[tpb]
   \vspace{1mm}
      \centering
      \includegraphics[width=0.45\textwidth]{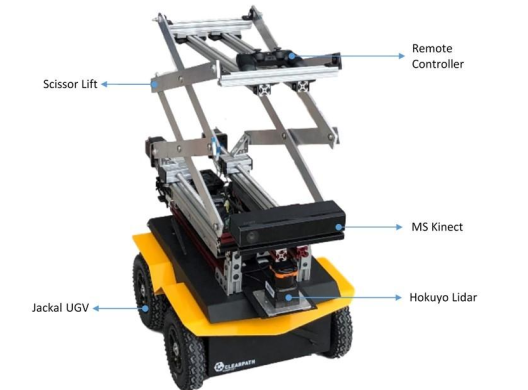}
      \caption{The robot used in the IAEA Robotics Challenge and experiments.}
      \label{robotconfig}
      \vspace{-1mm}
   \end{figure}

To meet the challenge requirements, we selected a small mobile robot as our development platform - Clearpath Jackal with 4x4 drivetrain for rugged all-terrain operation. We equipped Jackal with a Microsoft Kinect v2 RGBD camera, a Hokuyo scanning rangefinder and a custom design scissor lift (Figure~\ref{robotconfig}). The RGBD camera is primarily used for cylinder identification, localization and pose estimation while the laser rangefinder is used for 2D SLAM. To perform gamma measurements, a mechanism that can raise the instrument to desired altitude was needed. We designed and prototyped a scissor lift actuated by two stepper motors for this purpose. The scissor lift has a lift range of 1~m with a maximum payload of 10kg, meeting IAEA's requirement.

We will now present our algorithm development approach here to meet the IAEA Robotics Challenge requirements to generate and update the rich information map. Although we focus on the UF6 cylinder detection specifically, a similar approach can be used for a variety of objects that can be found in warehouses or other similar environments. The main contribution of our work is the autonomous navigation framework for efficiently generating and updating the rich information map in nuclear environments for inspection purposes. We also present experimental results to validate our approach. 


The paper is organized as follows: Section II presents the related literature. Section III describes our methodology for generating a rich information map. In Section IV, we present experimental evaluation and results in our lab environment. Lastly, Section V includes our conclusions with future research directions.

\section{RELATED WORK}

Numerous approaches can be found in literature addressing semantic mapping. In most cases, a simultaneous localization and mapping (SLAM) method is unified with object detection. That is, while mapping the environment on the background, the robot is actively searching for registered objects and localize detected objects on the map. Various object detection techniques are implemented to achieve the goal. In the 2D domain of object detection, a robot can generate a semantic map for a hallway by recognizing door signs when manually driven\cite{6095152}. \cite{4058387} realizes a semantic SLAM for more general objects using features from 2D images. \cite{6215666} achieves a relatively better object recognition results by combining 3D point clouds and 2D features. However, the methods proposed above only estimate the location of the target object without orientation. Instead, \cite{8265293} predicts both the location and the pose of objects by registering 2D feature descriptors to 3D voxel in the object frame. Instead of generating a map with discrete object information, \cite{7989538} uses Convolutional Neural Networks (CNNs) for prediction and generates a map labelling the entire scene, which also includes walls, floors, doors, etc.. Another approach for generating a rich information map is converting a set of existing individual datasets to one map that contains information about the objects offline \cite{Rusu2010}. The approach we propose in this paper is different from related works in two aspects. First, the object detection method applies a fully-convolutional model on the whole image \cite{2016arXiv161208242R}. On one hand, it has a better generalizability and efficiency than methods based on features (like SIFT). On the other hand, it has the advantage of informing the predictions with global context of the entire image. Second, we also investigate how the robot can generate a rich information map with all objects of interest detected autonomously, which is necessary for a fully autonomous inspection system.

In addition to generating a semantic map, the fusion of SLAM and object detection has many other applications. Object recognition can be enhanced by incorporating multiple viewpoint detection with the support from SLAM \cite{2015arXiv150601732P}. For instance, \cite{4058780} uses object detection to achieve a robust SLAM in home environments by treating detected objects as landmarks. Object detection can also improve SLAM quality by detecting and removing occlusion such as walking people in the static environment \cite{8098697}. \cite{8024025} takes advantage of object detection to assist a blind person without generating a map. They detect structural objects such as stairways or doorways and use the detected results to navigate a blind person to a destination.

\section{GENERATION OF RICH INFORMATION MAP}
   \begin{figure}[tpb]
      \centering
      \includegraphics[width=0.475\textwidth]{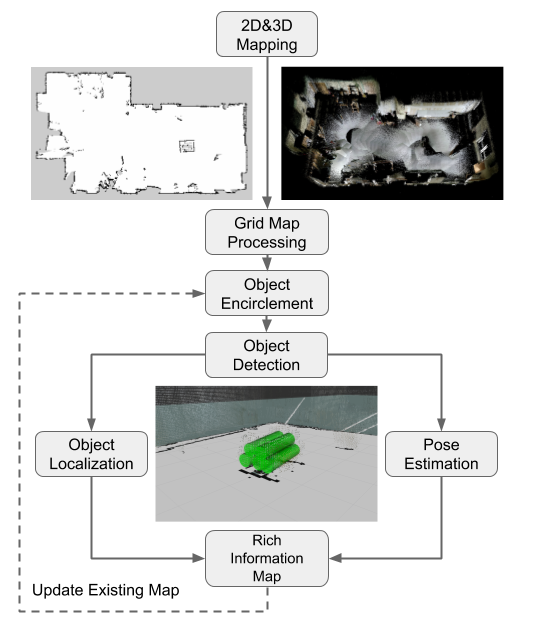}
      \caption{The flowchart of generating a rich information map.}
      \label{map}
   \end{figure}
In order to record all cylinder piles in the workspace without omission, we generate the rich information map in two steps: we first generate a 2D occupancy grid map of the entire workspace and then enrich the grid map. By doing this, we can perform cylinder detection from a global perspective, i.e., we can estimate the regions cylinders are located before detection happens. This is preferred because the search area is reduced and the critical region receives special focus. The process flow of generating the rich information map is shown in Figure \ref{map}. As you may note, we also generated a 3D point cloud map which aims to meet the challenge requirement. An active SLAM method is at the core of the navigation framework. In our experiment, we use the open source SLAM method RTAB-Map \cite{Labbé2017} together with frontier based exploration \cite{613851} to achieve navigation in an unknown environment. RTAB-Map is a graph-based method that is real time capable and supporting various sensor configurations. The frontier based active SLAM provides the robot a grid map with information classifying free and occupied areas, which is used for path planning. The robot then processes the generated grid map and identifies occupied regions that potentially contain cylinders. After that, the robot pursues towards the identified occupied area to seek for the cylinders that need to be inspected. Once the robot is near the cylinder pile, it starts counting and localizing UF6 cylinders and mark detected cylinders on the map. To avoid duplicate marking, the position of every detected cylinder is compared with the saved ones. The newly detected cylinder is considered as valid only if its distance to existing cylinders is greater than a prespecified threshold, denoted by \textalpha. The final generated map consists of two parts: the grip map registering free and occupied spaces and a database containing the cylinder information, such as position and orientation. The method also introduces an update mode to reflect cylinder number changes in a pile without regenerating a new map. In view of this general concept of operations, we will now discuss the details of the process.

\subsection{Grid Map Processing}

\begin{figure}[tb]
      \centering
      \subfigure[Original Map]{
      \includegraphics[width=0.22\textwidth]{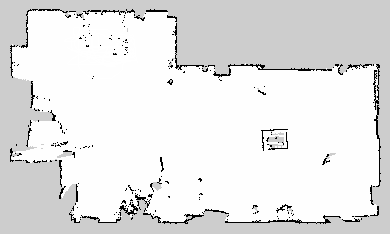}
      \label{grida}}
      \subfigure[Map after closing operation]{
      \includegraphics[width=0.22\textwidth]{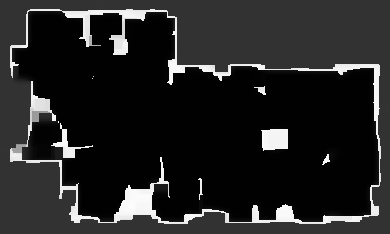}
      \label{gridb}}
      \subfigure[Contours and convex hulls]{
      \includegraphics[width=0.22\textwidth]{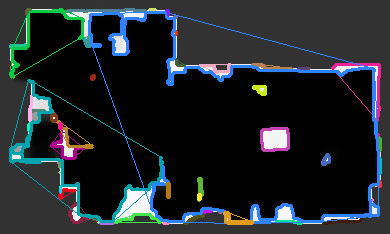}
      \label{gridc}}
      \subfigure[Filtered obstacle region]{
      \includegraphics[width=0.22\textwidth]{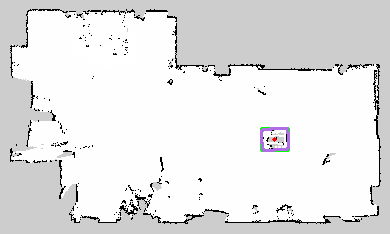}
      \label{gridd}}
      \caption{Grid map processing procedures and corresponding results.}
      \label{gridmap}
    \end{figure}

Detecting an object in an unknown environment can be time-consuming and, depending on the application, may require high detection rate when the robot is randomly exploring the area. The goal of the grid map processing is to identify the occupied regions that potentially contain cylinders from the occupancy grid map. This results in a reduced area for the robot to navigate and perform the detection in the area. 
We convert the generated grid map from SLAM to an image and utilize OpenCV to achieve occupied area classification. Figure \ref{gridmap} shows the grid map processing procedures and the results. For the map image, we first invert colors so that white regions represent obstacles and black regions are freespace. We do this because it is easier to find the contours for white regions when using OpenCV. We then perform dilation followed by erosion (known as closing operation) on the inverted image to reduce noise (Figure \ref{gridb}). After this step, all contours in the map image and their individual convex hulls are calculated (Figure \ref{gridc}). In order to obtain a clearer result, a filtering process is used. We first remove all open contours, because after the closing operation the cylinder pile region should be enclosed by one contour. For remaining closed contours, only the ones with an area that is greater than the projection area of a cylinder are kept. The location of filtered obstacle region can then be calculated and used for navigation. It should be noted that (Figure \ref{gridd}) corresponds to a case in which only one region is identified corresponding to the pile. However, grid map processing may result in more than one region for potential cylinder piles. Since these identified obstacle regions are used to navigate the robot to potential cylinder piles, only the true cylinder pile locations will eventually be saved.

\subsection{Object Encirclement}

To perform cylinder detection in the occupied region identified by grid map processing, we implement an object encirclement method assuming objects are stored in the form of piles, which is the case for UF6 cylinders. With the object encirclement, the pile of cylinders are treated as one large object for the robot to patrol around. The main idea behind grid map processing and object encirclement is to reduce  the search area and pre-define the search trajectory for the robot. 

   \begin{figure}[tpb]
   \vspace{3mm}
      \centering
      \includegraphics[width=0.45\textwidth]{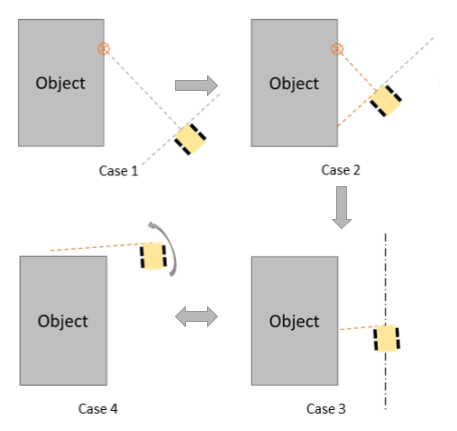}
      \caption{The illustration of the object encirclement algorithm.}
      \label{encirclement}
      \vspace{-3mm}
   \end{figure}

Figure \ref{encirclement} illustrates the process for the robot encircling the pile of cylinders. There are 4 $Phases$ to consider when the robot is executing an encirclement. In $Phase \ 1$, the desired location from grid map processing is sent to the robot. When the robot is close to the desired location, it starts calculating its distance from the object using laser scan information. Once the distance between the robot and the object is less than the threshold, \textbeta, the process transitions to $Phase \ 2$. In $Phase \ 2$, the robot estimates whether the object is to its right or left using the distance readings from the right and left quarters. The object is assigned to the side with a smaller distance. Once the object direction is identified by the robot, it remains unchanged and the process progresses to $Phase \ 3$. In $Phase \ 3$, a PD controller is implemented to ensure the robot is maintaining a constant distance, \textgamma, from the side of object. $Phase \ 4$ is triggered when the robot completes traversing the side of the object. This can be detected when the side distance from the object suddenly increases. A 90$^\circ$ turn is then performed in the direction of the pile and the encirclement continues.

By switching between $Phase \ 3$ and $Phase \ 4$, the robot can successfully achieve the encirclement of the pile of cylinders. During the object encirclement, the robot stops and performs a 90$^\circ$ turn to face the pile of cylinders every second. This is to detect the ID plate and perform the cylinder localization and pose estimation if an ID plate is found. The robot will exit the encirclement mode if no new cylinders are found for eight continuous 90$^\circ$ turns, which is equivalent to two encirclement. The robot can thus efficiently find all UF6 cylinders in the same pile since the search area is greatly reduced.

\subsection{Object Localization and Pose Estimation}

\begin{figure}[tbp]
\vspace{3mm}
      \centering
      {\includegraphics[width=0.22\textwidth]{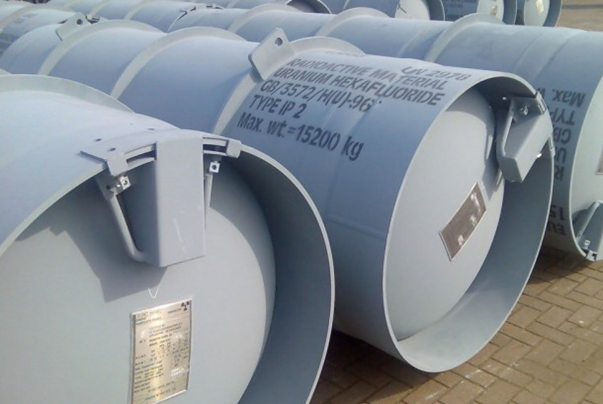}}
      {\includegraphics[width=0.22\textwidth]{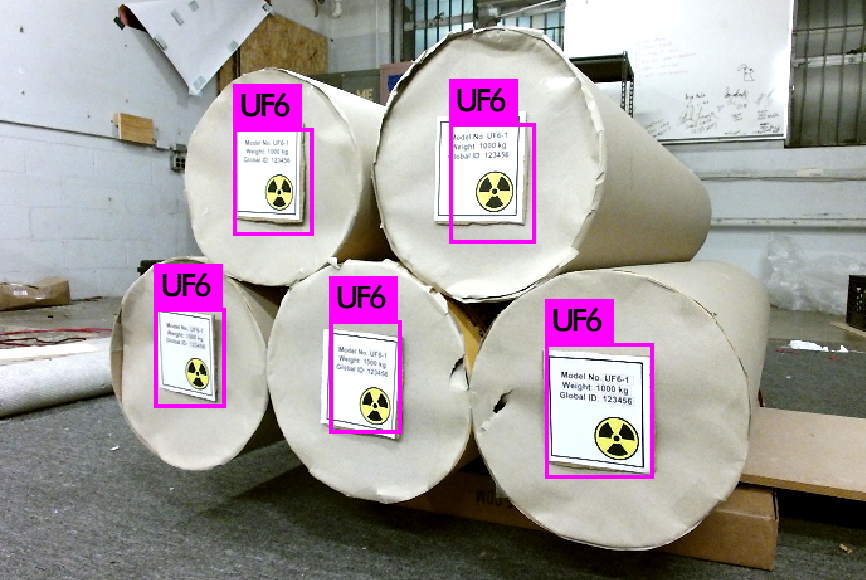}}
      \caption{The ID plate needs to be recognized. Left figure shows the ID plate on real UF6 cylinders. Right figure shows the mock up cylinder and detection result during our experiments.}
      \label{tag}
      \vspace{-3mm}
\end{figure}

To localize the objects of interest, we first apply a real time object detection system named YOLO to recognize the object from RGB image \cite{2016arXiv161208242R}. As one of the state-of-the-art methods, YOLO has the advantage of predicting with global context information and high speed allowing implementation for practical applications. For the UF6 cylinders, the goal is to identify the ID plate attached to one base of the cylinder. Figure \ref{tag} shows the ID tags on real UF6 cylinders and mock up UF6 cylinders used during our experiments.

Once the ID plate is detected in the image, its central location in the camera frame can be calculated using the perspective projection equation:
\begin{equation}
    x=f*\frac{X}{Z}, \quad y=f*\frac{Y}{Z}
    \label{eq1}
\end{equation}
where $f$ is the focal length of the camera, $(x, y)$ are coordinates of the detected object in 2D image plane, which is an output of YOLO and $(X, Y, Z)$ are the coordinates of the detected object in the camera frame. $Z$, the depth, is read directly and $X, Y$ are calculated using \ref{eq1}. By applying a frame transformation, the location of the detected object in the map frame is calculated.

To mark the cylinder on the map, the orientation of the cylinder is also needed. Since the base of the cylinder is a planar surface, the orientation of the cylinder can be estimated by calculating its normal vector. To do so, we first extract point clouds within a cube of 0.5~m centered at the calculated location of the ID plate. These extracted point clouds mainly come from the base surface of the cylinder. Then a planar segmentation using RANSAC on extracted point clouds is performed to find a vertical plane. The normal vector to this vertical plane then represents the axial direction of the UF6 cylinder. As the dimensions of standard UF6 cylinders are known, the cylinder with its known position and orientation can be marked on the map.

\subsection{Map Update}

The generated rich information map obtained by the process described above can be saved for performing repetitive inspection tasks. However, one challenge of using an existing map is that the number of stored UF6 cylinders may vary between inspections. With the changing number of UF6 cylinders, the rich information map generated during an earlier inspection will not provide correct information for the robot to perform a new inspection. To solve this problem, a method to update the map is required as the number of cylinders in each pile varies. It is assumed that the number of cylinders in each pile may vary but the locations of cylinder piles are not changed, which is the case in the IAEA Robotics Challenge. 

First, a recheck action is performed by the robot after loading the information about the saved cylinders from a prior inspection. For cylinders in the same pile, the robot will navigate to the pile and perform an object detection as explained earlier. If a cylinder is not found at the registered location, the robot will delete the saved cylinder(s) from the database. Cylinders that are found at their registered locations will remain in the database. Once all saved cylinders in one pile are verified, the object encirclement process begins to detect if new cylinders are added in to the pile. By doing the recheck for every pile, the robot is able to keep the rich information map updated without exploring the whole environment from scratch. This map updating mode is only needed when the count of stored cylinders is changed, and the robot should be informed ahead by the human inspectors. In other cases, the robot can use the map as a guide directly and navigate to registered cylinder locations to perform the inspection.

\section{EXPERIMENTAL VALIDATION}

\begin{figure}[t]
\vspace{3mm}
      \centering
      {\includegraphics[width=0.22\textwidth]{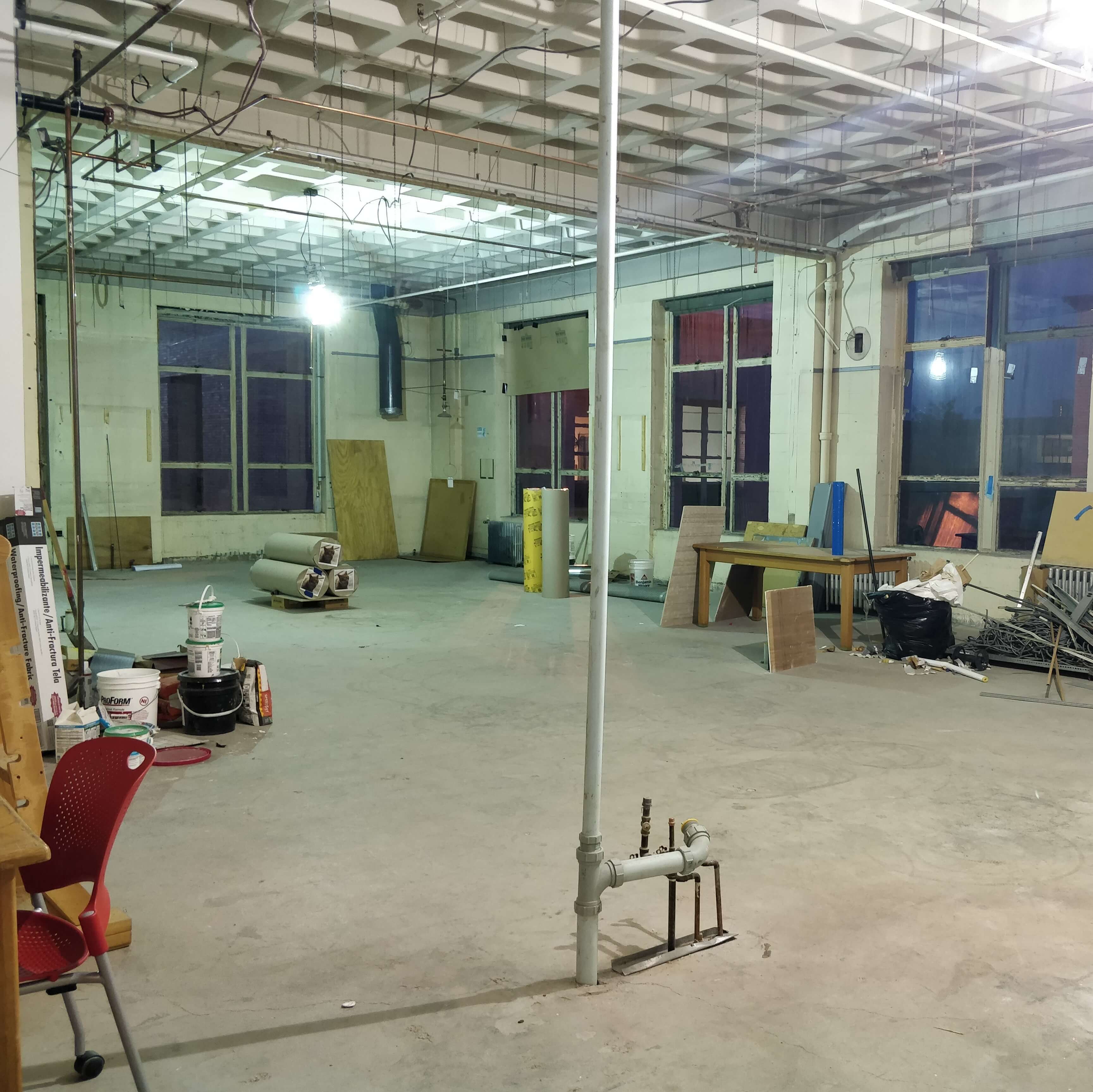}}
      {\includegraphics[width=0.22\textwidth]{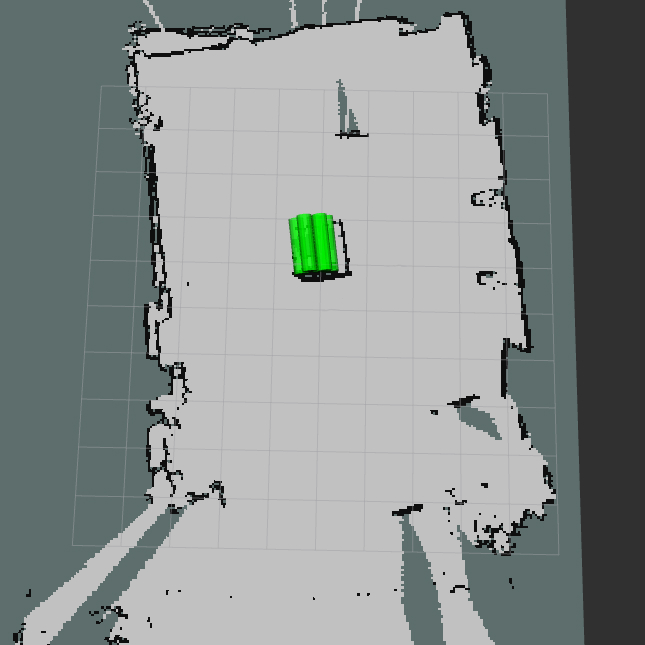}}
      \caption{The experimental environment and the map generated. The green models on the map shown in right figure represent the detected UF6 cylinder mock ups.}
      \label{env}
      \vspace{-3mm}
\end{figure}

To evaluate the effectiveness of the proposed inspection platform, experiments in a simulated testbed environment with mock ups of UF6 cylinder were performed. Five UF6 cylinder mock ups, 0.3$m$ in diameter and 1.2$m$ long, were built using cardboard tubes. Fiducial markers are placed on each cylinder to represent the ID tags. We trained a model to detect the ID plate with Tiny Yolo \cite{2016arXiv161208242R} and the detection results are shown in Figure \ref{tag}.

We designed two experiments to evaluate the performance of the system on new map generation and existing map updates. Figure \ref{env} depicts the experimental environment and the corresponding map. To make the map readable by human inspectors, cylinders found using 3D models are marked on the 2D map.

\subsection{Experiment 1: Search and Map Generation}

\begin{table}[b]
\centering
\begin{tabular}{|c|c|c|c|}
\hline
Trial & Cylinders Found &  Cylinders Localized & Pose Estimated \\ \hline
1                & 5/5             & 5/5                 & 5/5             \\ \hline
2                & 5/5             & 3/5                 & 5/5             \\ \hline
3                & 5/5             & 2/5                 & 5/5             \\ \hline
\end{tabular}
\caption{Results of 3 trials of Experiment 1.}
\label{tab:t1}
\end{table}

\begin{figure}[t]
\vspace{3mm}
      \centering
      {\includegraphics[width=0.23\textwidth]{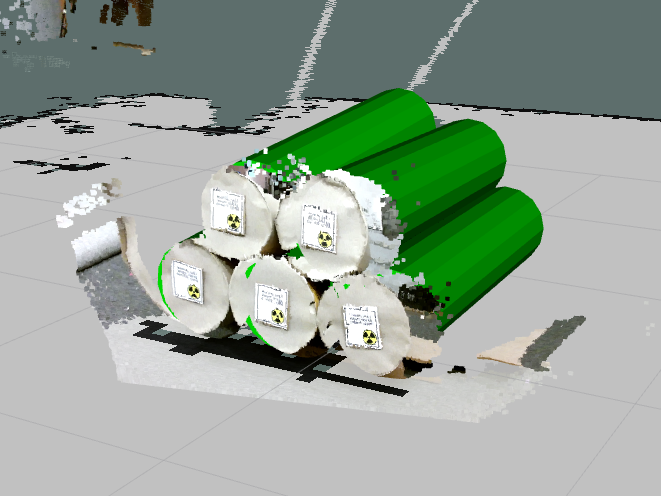}}
      {\includegraphics[width=0.23\textwidth]{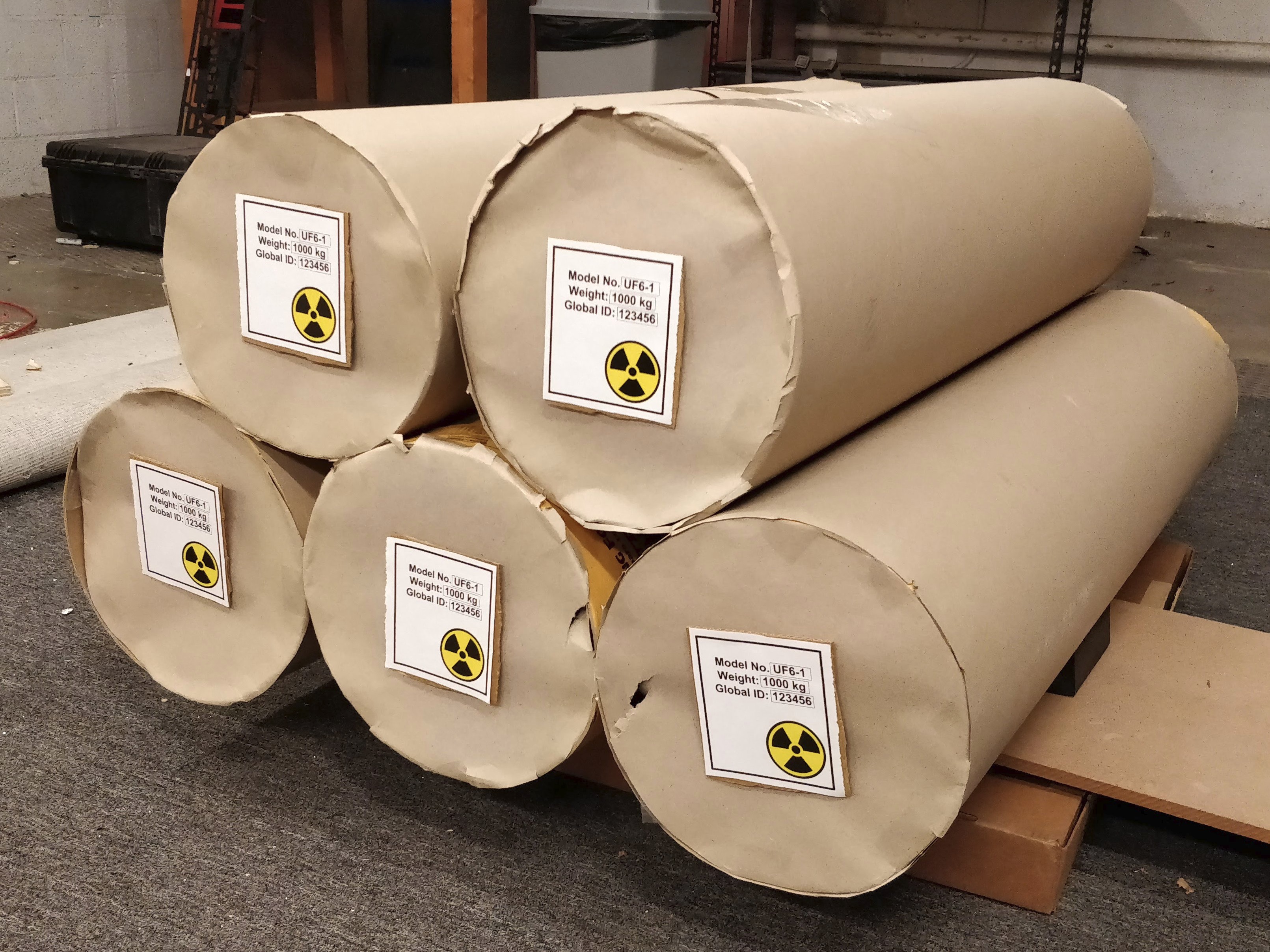}}
      \caption{Results of experiment 1, (left) cylinder model with estimated pose and location on the generated map, (right) the actual pile in the environment.}
      \label{e1}
      \vspace{-2mm}
\end{figure}

The goal of experiment 1 is to test the robot's ability to map an unknown environment and mark all 5 mock ups on the map correctly. We stored the 5 mock ups in one pile at a central area in the environment and the robot is initially located at the entrance to the room. The experiments were repeated three times using the process presented above. Results for 3 trials are presented in Table~\ref{tab:t1}. It is observed that the robot generated an acceptable map of the room and identified all 5 cylinders, during the 3 trials.  In one trial,all 5 cylinders have been localized accurately, however there were location offsets and overlapping of cylinders were observed in 2 trials. The axial direction of all 5 cylinders were accurately calculated for all 3 trials using the proposed orientation estimation technique. Figure \ref{e1} depicts the results from the experiment in which all 5 cylinders are localized accurately. The cylinder model from the detection is matched to the point cloud.

\subsection{Experiment 2: Map Update}

\begin{figure}[t]
      \centering
      {\includegraphics[width=0.23\textwidth]{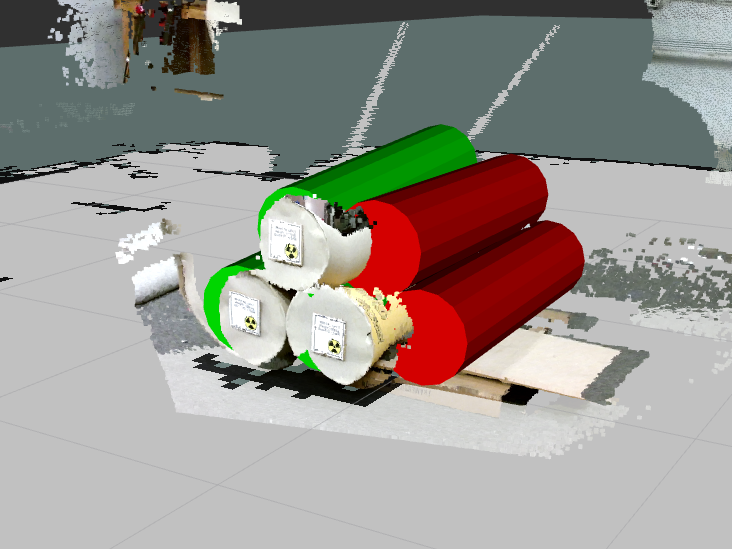}}
      {\includegraphics[width=0.23\textwidth]{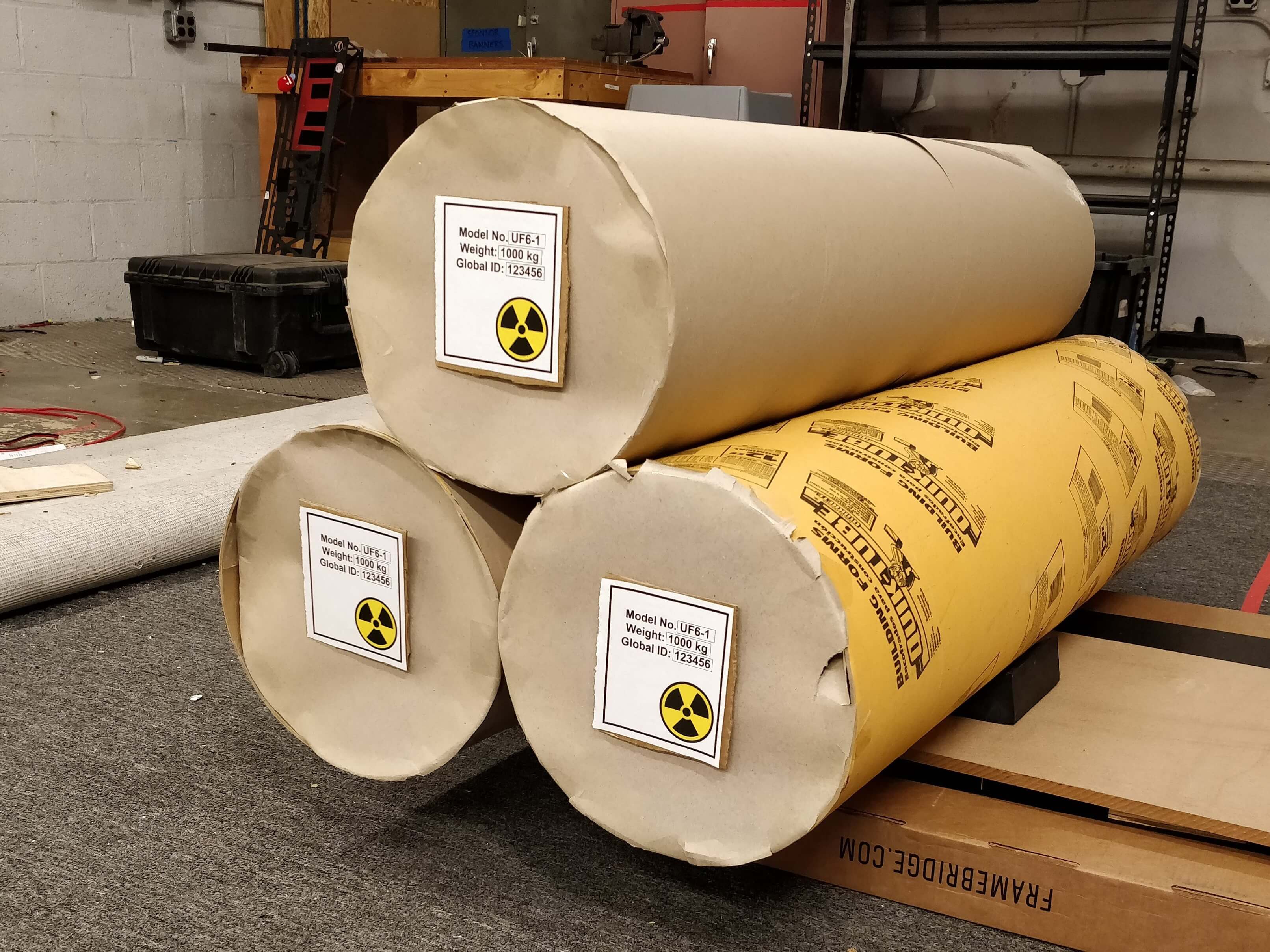}}
      \caption{Result of experiment 2, (left) red indicates the missing cylinders at the pre-registered locations, green indicates the cylinder are found.}
      \label{e2}
\end{figure}

The goal of experiment 2 is to test the effectiveness of our approach in updating an existing map to reflect the changes to the stored cylinders. For the case of cylinders added to a pile, the robot will just perform another object encirclement to register the new cylinders. We, thus mainly focus on testing the case when cylinders are removed from a pile. During the experiment, we removed 2 cylinders from the pile and launch the robot with database of 5 localized cylinders from experiment 1. This experiment was also performed 3 times. Of all three trials,  2 missing cylinders were detected correctly and removed from the database. Figure \ref{e2} shows the results from experiment 2 and visualization for the missing cylinders to update the map. The green cylinders are the ones found at the recorded positions while the red cylinders indicate the ones missing from the pile.

\subsection{Discussion}

The experimental results validate that the proposed method to perform inspections in nuclear storage facilities. With the object encirclement algorithm, the robot is able to find and localize target cylinders and generate a rich information map. Later, the cylinder database can be updated to reflect the changes. However, the experiments also revealed a number of issues for future development.

First, the localization offset can be an obstacle in generating an accurate map. The results of experiment 1 show that even the robot can find all cylinders, it might not be able to localize them all correctly. The are two kinds of localization offsets, namely along the radial and axial directions. The radial offset comes from the object detection system which can detect the object without seeing the entire picture (the ID plate). This robust feature ensures that the robot can find all cylinders in a pile. However, this also provides a bias in the center position when the system outputs the detection results from a partial image. As for the axial offset, it mainly comes from an inaccurate depth measurement by the RGBD camera. To solve the localization offset problem, a compliance test between all detected cylinders is needed. 

Second, the object encirclement can be further optimized to increase the object marking efficiency. Since the ID plates are all placed at the same side of the pile, it is unnecessary for the robot to keep encircling the whole pile once the robot determined the side where ID plates locate. The robot can switch to patrol along one side of the pile to speed up the process.

\section{CONCLUSIONS}

We presented a method to generate and update a rich information map for inspections in nuclear storage facilities by autonomous robots. By unifying SLAM and object detection, a world map representing not only free/occupied locations but also position and orientation of the stored UF6 cylinders is generated. Considering the fact that UF6 cylinders are usually stored in the form of piles, we proposed an object encirclement method for detecting cylinders effectively. The generated rich information map is used both for robot navigation and as a visual tool for human inspectors to monitor UF6 container inventory. 

To evaluate the performance of the proposed method, we performed two sets of experiments. We can conclude from the results of experiment 1 that the robot is capable of identifying and locating cylinders. The performance regarding localizing the cylinders is functional but can be improved in the future. The results from experiment 2 confirm that the robot is able to update an existing map to reflect changes in the cylinder storage.

The localization offset is one obstacle for achieving a higher accuracy in rich information mapping and we plan to work on a compliance test between detected cylinders to improve the localization precision. Furthermore, object encirclement method will be optimized to improve mapping efficiency for an extended number of application scenarios for inspection in nuclear environments.

\addtolength{\textheight}{-12cm}   



\section*{ACKNOWLEDGMENT}

This material is based upon work supported by the Department of Energy, Office of Environmental Management under Grant No. DE-EM0004482. The authors would also like to acknowledge the support provided by the United States Support Program to IAEA Safeguards under a contract issued by Brookhaven Science Associates/Brookhaven National Laboratory for a two person team from the Robotics and Intelligence Vehicle Research Laboratory at Northeastern University to travel to and participate in the second stage of the IAEA Robotics Challenge held in Pullenvale, Australia.


\bibliographystyle{IEEEtran}
\bibliography{citation}

\begin{thebibliography}{10}
\providecommand{\url}[1]{#1}
\csname url@samestyle\endcsname
\providecommand{\newblock}{\relax}
\providecommand{\bibinfo}[2]{#2}
\providecommand{\BIBentrySTDinterwordspacing}{\spaceskip=0pt\relax}
\providecommand{\BIBentryALTinterwordstretchfactor}{4}
\providecommand{\BIBentryALTinterwordspacing}{\spaceskip=\fontdimen2\font plus
\BIBentryALTinterwordstretchfactor\fontdimen3\font minus
  \fontdimen4\font\relax}
\providecommand{\BIBforeignlanguage}[2]{{%
\expandafter\ifx\csname l@#1\endcsname\relax
\typeout{** WARNING: IEEEtran.bst: No hyphenation pattern has been}%
\typeout{** loaded for the language `#1'. Using the pattern for}%
\typeout{** the default language instead.}%
\else
\language=\csname l@#1\endcsname
\fi
#2}}
\providecommand{\BIBdecl}{\relax}
\BIBdecl

\bibitem{Labbé2017}
\BIBentryALTinterwordspacing
M.~Labb{\'e} and F.~Michaud, ``Long-term online multi-session graph-based splam
  with memory management,'' \emph{Autonomous Robots}, Nov 2017. [Online].
  Available: \url{https://doi.org/10.1007/s10514-017-9682-5}
\BIBentrySTDinterwordspacing

\bibitem{7946260}
R.~Mur-Artal and J.~D. Tardós, ``Orb-slam2: An open-source slam system for
  monocular, stereo, and rgb-d cameras,'' \emph{IEEE Transactions on Robotics},
  vol.~33, no.~5, pp. 1255--1262, Oct 2017.

\bibitem{KohlbrecherMeyerStrykKlingaufFlexibleSlamSystem2011}
S.~Kohlbrecher, J.~Meyer, O.~von Stryk, and U.~Klingauf, ``A flexible and
  scalable slam system with full 3d motion estimation,'' in \emph{Proc. IEEE
  International Symposium on Safety, Security and Rescue Robotics
  (SSRR)}.\hskip 1em plus 0.5em minus 0.4em\relax IEEE, November 2011.

\bibitem{8265293}
\BIBentryALTinterwordspacing
D.~D. Gregorio, T.~Cavallari, and L.~D. Stefano, ``Skimap++: Real-time mapping
  and object recognition for robotics,'' in \emph{2017 IEEE International
  Conference on Computer Vision Workshop (ICCVW)}, vol.~00, Oct. 2017, pp.
  660--668. [Online]. Available:
  \url{doi.ieeecomputersociety.org/10.1109/ICCVW.2017.84}
\BIBentrySTDinterwordspacing

\bibitem{6095152}
J.~G. Rogers, A.~J.~B. Trevor, C.~Nieto-Granda, and H.~I. Christensen,
  ``Simultaneous localization and mapping with learned object recognition and
  semantic data association,'' in \emph{2011 IEEE/RSJ International Conference
  on Intelligent Robots and Systems}, Sept 2011, pp. 1264--1270.

\bibitem{4058387}
S.~Ekvall, P.~Jensfelt, and D.~Kragic, ``Integrating active mobile robot object
  recognition and slam in natural environments,'' in \emph{2006 IEEE/RSJ
  International Conference on Intelligent Robots and Systems}, Oct 2006, pp.
  5792--5797.

\bibitem{6215666}
D.~Filliat, E.~Battesti, S.~Bazeille, G.~Duceux, A.~Gepperth, L.~Harrath,
  I.~Jebari, R.~Pereira, A.~Tapus, C.~Meyer, S.~H. Ieng, R.~Benosman,
  E.~Cizeron, J.~C. Mamanna, and B.~Pothier, ``Rgbd object recognition and
  visual texture classification for indoor semantic mapping,'' in \emph{2012
  IEEE International Conference on Technologies for Practical Robot
  Applications (TePRA)}, April 2012, pp. 127--132.

\bibitem{7989538}
J.~McCormac, A.~Handa, A.~Davison, and S.~Leutenegger, ``Semanticfusion: Dense
  3d semantic mapping with convolutional neural networks,'' in \emph{2017 IEEE
  International Conference on Robotics and Automation (ICRA)}, May 2017, pp.
  4628--4635.

\bibitem{Rusu2010}
\BIBentryALTinterwordspacing
R.~B. Rusu, ``Semantic 3d object maps for everyday manipulation in human living
  environments,'' \emph{KI - K{\"u}nstliche Intelligenz}, vol.~24, no.~4, pp.
  345--348, Nov 2010. [Online]. Available:
  \url{https://doi.org/10.1007/s13218-010-0059-6}
\BIBentrySTDinterwordspacing

\bibitem{2016arXiv161208242R}
J.~{Redmon} and A.~{Farhadi}, ``{YOLO9000: Better, Faster, Stronger},''
  \emph{ArXiv e-prints}, Dec. 2016.

\bibitem{2015arXiv150601732P}
S.~{Pillai} and J.~{Leonard}, ``{Monocular SLAM Supported Object
  Recognition},'' \emph{ArXiv e-prints}, Jun. 2015.

\bibitem{4058780}
S.~Ahn, M.~Choi, J.~Choi, and W.~K. Chung, ``Data association using visual
  object recognition for ekf-slam in home environment,'' in \emph{2006 IEEE/RSJ
  International Conference on Intelligent Robots and Systems}, Oct 2006, pp.
  2588--2594.

\bibitem{8098697}
L.~Riazuelo, L.~Montano, and J.~M.~M. Montiel, ``Semantic visual slam in
  populated environments,'' in \emph{2017 European Conference on Mobile Robots
  (ECMR)}, Sept 2017, pp. 1--7.

\bibitem{8024025}
C.~Ye and X.~Qian, ``3-d object recognition of a robotic navigation aid for the
  visually impaired,'' \emph{IEEE Transactions on Neural Systems and
  Rehabilitation Engineering}, vol.~26, no.~2, pp. 441--450, Feb 2018.

\bibitem{613851}
B.~Yamauchi, ``A frontier-based approach for autonomous exploration,'' in
  \emph{Computational Intelligence in Robotics and Automation, 1997. CIRA'97.,
  Proceedings., 1997 IEEE International Symposium on}, Jul 1997, pp. 146--151.

\end{thebibliography}

\end{document}